\documentclass[runningheads]{llncs}
\usepackage{graphicx}
\usepackage{tikz}
\usepackage{comment}
\usepackage{amsmath,amssymb} % define this before the line numbering.
\usepackage{color}
\usepackage{multirow}
\usepackage[accsupp]{axessibility}

\usepackage{booktabs}
\usepackage{algpseudocode}
\usepackage{algorithm}
\usepackage{algorithmicx}
\usepackage{setspace}
\usepackage{stfloats}
\usepackage{colortbl}
\usepackage{marvosym}
\usepackage[colorlinks = true,
            linkcolor = red,
            urlcolor  = magenta,
            citecolor = green,
            anchorcolor = blue]{hyperref}
\begin{document}
\pagestyle{headings}
\mainmatter
\def\ECCVSubNumber{7687}  % Insert your submission number here

\title{FRT-PAD: Effective Presentation Attack Detection Driven by Face Related Task}
\author{Wentian Zhang\inst{\dagger \; 1,2} \and Haozhe Liu\inst{\dagger \; 3} \and Feng Liu\inst{\textrm{\Letter} \; 1,2} \and Raghavendra Ramachandra\inst{4} \and Christoph Busch \inst{4}}

\authorrunning{W. Zhang et al.}
\titlerunning{Effective Presentation Attack Detection Driven by Face Related Task}

\institute{Computer Vision Institute, Shenzhen University, Shenzhen, China \and
Institute of Artificial Intelligence and Robotics for Society, Shenzhen, China 
\and 
King Abdullah University of Science and Technology, Saudi Arabia
\and Norwegian Biometrics Laboratory (NBL), Norwegian University of Science and Technology, Gjøvik 2818, Norway \\
\email{zhangwentianml@gmail.com; haozhe.liu@kaust.edu.sa; feng.liu@szu.edu.cn; \{raghavendra.ramachandra; christoph.busch\}@ntnu.no} \\
\url{https://github.com/WentianZhang-ML/FRT-PAD}}

\footnotetext{$\dagger$ Equal Contribution}
\footnotetext{This work was done when Haozhe Liu was a visiting student at NTNU, Norway, under the supervision of R. Ramanchandra and C. Busch.}
\footnotetext{This paper is accepted by ECCV'2022.}
\maketitle

\begin{abstract}
The robustness and generalization ability of Presentation Attack Detection (PAD) methods is critical to ensure the security of Face Recognition Systems (FRSs). However, in a real scenario, Presentation Attacks (PAs) are various and it is hard to predict the Presentation Attack Instrument (PAI) species that will be used by the attacker. Existing PAD methods are highly dependent on the limited training set and cannot generalize well to unknown PAI species. Unlike this specific PAD task, other face related tasks trained by huge amount of real faces (e.g. face recognition and attribute editing) can be effectively adopted into different application scenarios. Inspired by this, we propose to trade position of PAD and face related work in a face system and apply the free acquired prior knowledge from face related tasks to solve face PAD, so as to improve the generalization ability in detecting PAs. The proposed method, first introduces task specific features from other face related task, then, we design a Cross-Modal Adapter using a Graph Attention Network (GAT) to re-map such features to adapt to PAD task. Finally, face PAD is achieved by using the hierarchical features from a CNN-based PA detector and the re-mapped features. The experimental results show that the proposed method can achieve significant improvements in the complicated and hybrid datasets, when compared with the state-of-the-art methods. In particular, when training on the datasets OULU-NPU, CASIA-FASD, and Idiap Replay-Attack, we obtain HTER (Half Total Error Rate) of 5.48\% for the testing dataset MSU-MFSD, outperforming the baseline by 7.39\%.

\keywords{Face, Presentation Attack Detection, Graph Neural Network}
\end{abstract}

\section{Introduction}
\label{sec:intro}
Face Recognition Systems (FRSs) are widely deployed in authentication applications especially in access control and mobile phone unlocking in our daily life~\cite{Karasugi2020Face,Yu2020Fair}. However, recent studies~\cite{ramachandra2017presentation,jia2020survey} demonstrate the existing face recognition systems are lacking robustness, since they are easily spoofed by presentation attacks (PAs), such as photographs, video replays, low-cost artificial masks~\cite{boulkenafet2017oulu} and facial make-up attacks~\cite{drozdowski2021makeup}. 
Meanwhile, the face images can be easily obtained from social media, which seriously increases the risk of PAs. These issues raise wide concerns about the vulnerability of facial recognition technologies. Consequently, it is crucial to detect PAs to achieve robust and reliable FRS.

To tackle such challenge, many face PAD methods have been proposed, which can be divided into hardware and software based methods. 
Hardware based solutions~\cite{heusch2020deep,raghavendra2015presentation} generally employ specific sensors to acquire presentations with different image modalities to detect PAs. Although, these methods provide strong robustness, their applicability is still limited because of unsatisfying performance to new application scenarios and cost limitations. 
Software based algorithms usually explore the distinctive features between bona fides (live faces) and PAs, such as hand-crafted features~\cite{de2013can,wen2015face,patel2016secure,boulkenafet2016face} and deep features~\cite{yang2014learn,george2019deep}. Due to the advances of deep learning in recent years, deep feature based methods have been widely used in the community, since better performance can be achieved by adopting convolutional neural networks (CNNs)~\cite{jia2020single,wang2021rgb,george2021cross}.

However, such learning-based method might not obtain ideal generalization to different unseen attacks, since they often needs comparable bona fide and PA samples for training. In the real scenario, PAs with different materials and instruments are hard to collect, the PAD mechanisms are limited by the unbalanced training data \cite{liu2021one,liu2021fingerprint}. On the contrary, face related tasks (e.g. face, expression and attribute editing) possess strong generalization capability, since they are trained by millions of live faces cross genders, ages and ethnic groups from specific datasets. We argue that face PAD task should share some common patterns with other face related tasks, and the performance of PA detection might be boosted by the features adopted from such tasks.
\begin{figure*}[h]
  \centering
   \includegraphics[width=0.75\linewidth]{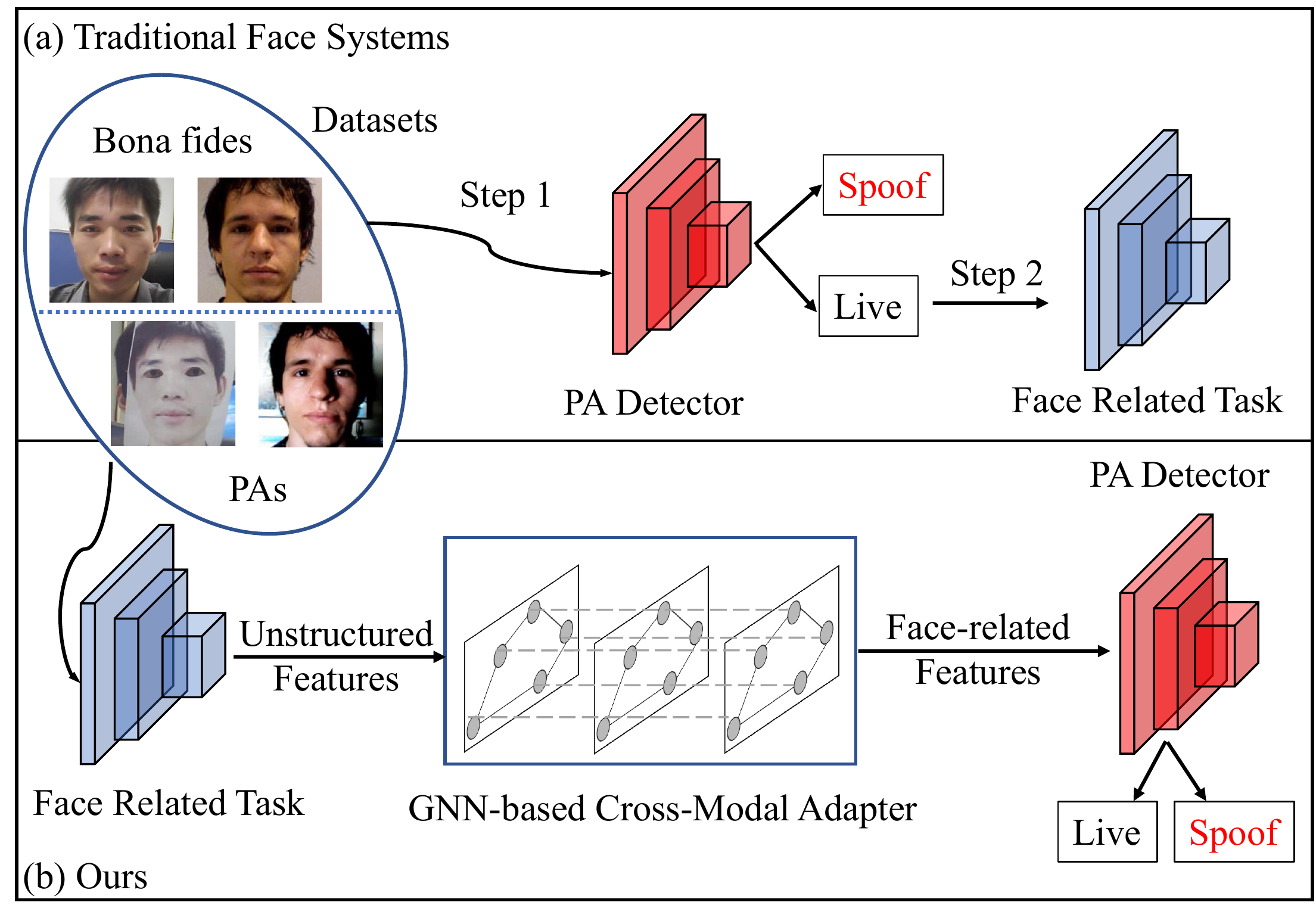}
   \caption{The diagram of (a) traditional face systems with PA detector and (b) the proposed scheme. In the proposed method, face systems, including face, expression and attribute editing, are directly implemented. We then apply a GNN-based Cross-Modal Adapter to adapt their extracted features, which contain abundant generalization knowledge, to facilitate PAD.}
   \label{fig:motivation}
\end{figure*}

As shown in Fig.~\ref{fig:motivation}(a), in traditional face systems, existing PAD mechanism like CNN-based PA detector, always plays a forward in face system and is independent with the following face related tasks. 
However, it is rarely discussed to contact PAD mechanism with face related tasks in the community. Since the face related tasks have already been trained, we consider that why not directly use these \textbf{free acquired} face-related features from face related tasks to serve for PAD. As shown in Fig.~\ref{fig:motivation}(b), different from traditional face systems, we attempt to directly implement the trained face related tasks in face systems, and then utilize their extracted task specific features to make PAD mechanism generalize well in real scenarios.

In this paper, we propose a PAD method utilizing feature-level prior knowledge from face related task, denoted as \textbf{FRT-PAD} to solve the above mentioned problem. 
First, we trade the position of PAD mechanism and face related task in a face system. The task specific features, which contain abundant generalization knowledge, are directly obtained from face related tasks. By following a Graph Neural Network (GNN), a Cross-Modal Adapter is put forward to re-map and adapt the task specific features to PAD task. The generalization capability is finally improved by alleviating the problem resulting in limited PA samples for training. The main contributions of this work are concluded as follows,
\begin{itemize}
  \item Existing PAD mechanism trained on the limited datasets are vulnerable to unseen PAs. To address this problem, we rethink the order of PAD mechanism and face related task in face system and directly introduce the free acquired features from face related tasks in PAD, which can improve generalization ability of PAD model.
  \item A Cross-Modal Adapter is designed to obtain face-related features, which can adapt task specific features into PAD space.
  \item The effectiveness and superiority of our method is evaluated on the public datasets. Particularly, when OULU-NPU~\cite{boulkenafet2017oulu}, CASIA-FASD~\cite{zhang2012face}, and Idiap Replay-Attack~\cite{chingovska2012effectiveness} are adopted as training set and MSU-MFSD~\cite{wen2015face} is used as test set, the HTER (Half Total Error Rate) of the proposed method can outperform the baseline by 7.39\%. 
\end{itemize}

%--------------------------------------------------------------------------------------------------------------------
\section{Related Works}
As face related task is for the first time introduced in face PAD, the proposed face PAD scheme is different from existing PAD methods. Hence, our reviews mainly include face PAD methods and face related tasks.

\subsection{Face Presentation Attack Detection}
Existing face PAD methods can be categorized into hand-crafted and deep learning based methods.
Hand-crafted methods employ the algorithms, e.g. LBP~\cite{de2013can}, IDA~\cite{wen2015face}, SIFT~\cite{patel2016secure}, and SURF~\cite{boulkenafet2016face} to extract the features and then adopt traditional classifiers such as LDA and SVM to detect PAs. However, the hand-crafted features can be easily influenced by the variations of imaging quality and illumination. As a result, feature based methods generally can not generalize well to different application scenarios. 

To address such challenges, deep learning models are then proposed for face PAD. Yang et al.~\cite{yang2014learn} proposed to use CNNs to extract deep discriminative features for face PAD. Nguyen et al.~\cite{nguyen2020attended} designed a multi-task learning model, which locates the most important regions of the input to detect PAs. 
Yu et al.~\cite{yu2020searching} proposed a Central Difference Convolution (CDC) structure to capture intrinsic detailed patterns for face PAD and then used Neural Architecture Search (NAS) in CDC based network to achieve a better result. 
Besides applying only RGB images, auxiliary information of face, e.g. face depth, are considered to establish a more robust detector. 
Liu et al.~\cite{liu2018learning} explored face depth as auxiliary information and estimated rPPG signal of RGB images through a CNN-RNN model for face PAD. 
George et al.~\cite{george2021cross} introduced a cross-modal loss function to supervise the multi-stream model, which extracted features from both RGB and depth channels. 
Liu et al.~\cite{liu2021taming} proposed a self-supervised learning based method to search the better initialization for face PAD. 
Although such deep learning methods can achieve better PAD performance, their dependence on training data would inevitably leads a bias when accessible data is limited. In particular, numerous studies \cite{george2020learning,shao2019multi,wang2020cross} have shown that, PA detectors trained on one dataset can not generalize to other datasets effectively.

To improve the generalization, researchers have further proposed one-class and domain generalization methods. 
A one-class multi-channel CNN model~\cite{george2020learning} was proposed to learn the discriminative representation for bona fides within a compact embedding space. 
Different from one-class methods, domain generalization based methods pay more attention to the disparities among the domains.
Shao et al.~\cite{shao2019multi} proposed a multi-adversarial deep domain generalization framework to learn a generalized feature space within the dual-force triplet-mining constraint. Since the learned feature space is discriminative and shared by multiple source domains, the generalization to new face PAs can be ensured effectively.  
Wang et al.~\cite{wang2020cross} disentangled PAD informative features from subject-driven features and then designed a multi-domain learning based network to learn domain-independent features cross different domains for face PAD. Generally speaking, when training data is adequate for the aforementioned method, the detector can achieve very competitive performance. However, unlike other face related tasks, Presentation Attacks (PAs) are hard to collect and the types/instruments of attacks are consistently increasing. Due to such open challenges, we propose a PAD mechanism based on face related task in a common face system to decrease the dependence of the PA detector on data scale. In particular, we obtain face-related features from face related tasks for a more robust representation with better generalization capability. Benefit from extensive samples collected in other tasks, face PA detector can achieve significant improvement within limited data.

\subsection{Face Related Tasks}
With the advances in deep learning, face related tasks including face recognition, face expression recognition, face attribute editing, etc, have become a very active field~\cite{deng2019arcface,wang2020suppressing,choi2018stargan}.
In this section, we briefly introduce some representative tasks adopted in this work due to the limited scope of the paper. For large-scale face recognition, Deng et al.~\cite{deng2019arcface} proposed an Additive Angular Margin Loss (ArcFace) , which has a clear geometric interpretation, to obtain highly discriminative features. ArcFace is a solid work evaluated on the various face recognition benchmarks, including image datasets with trillions of pairs and a large-scale video dataset.
In the terms of facial expression recognition, Wang et al.~\cite{wang2020suppressing} proposed a Self-Cure Network (SCN) to address uncertainties in facial expressions. By combining self-attention and relabelling mechanism, such method can prevent deep networks from over-fitting uncertain facial images. 
Choi1 et al.~\cite{choi2018stargan} proposed a scalable approach called StarGAN, to perform image-to-image translations for multiple domains and achieve facial attribute transfer. StarGAN can flexibly learn reliable features universally applicable to multiple domains of images with different facial attribute values, which is always set as a famous baseline in generative task. As the representative works for the corresponding tasks, the aforementioned solutions are conducted in this paper to investigate the relationship between PAD and face related tasks. 

\section{Proposed Method}
\begin{figure*}[h!]
  \centering
   \includegraphics[width=0.99\linewidth]{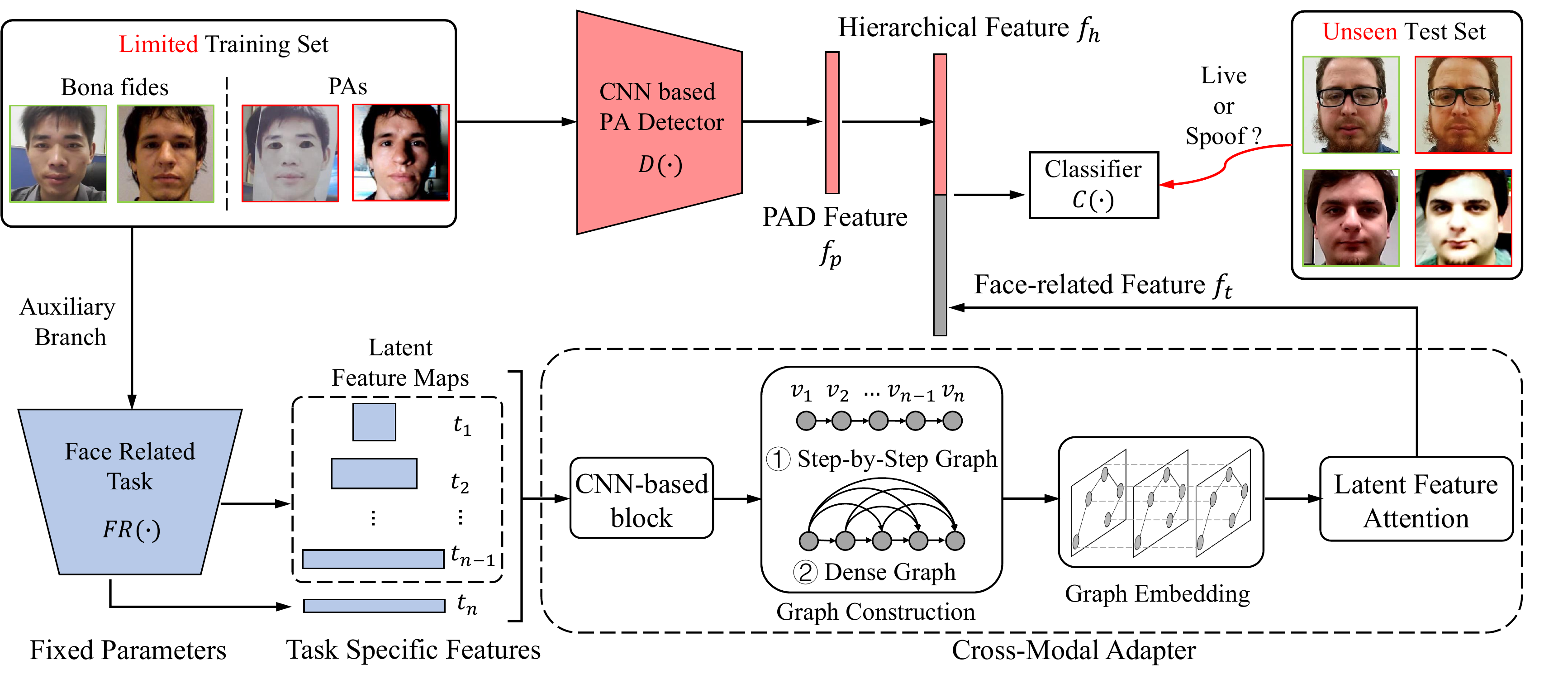}
   \caption{The pipeline of our proposed PAD mechanism based on face related task in a common face system (\textbf{FRT-PAD}). In auxiliary branch, the task specific features $t_i$ can be extracted by the parameter-fixed model $FR(\cdot)$, which has been trained from face related tasks. Then, a CNN-based block is used to transform $t_i$ to graph vertexes. We construct a Graph Attention Netwok (GAT) to re-map $t_i$ to fit PAD. By following latent feature attention, the face-related feature $f_t$ is obtained, and finally fused with the main branch to achieve face PAD. }
   \label{fig:pipeline}
\end{figure*}
In this paper, we propose a face-related-task based PAD mechanism (\textbf{FRT-PAD}) to improve the generalization capability of PAD model.
As shown in Fig.~\ref{fig:pipeline}, the proposed \textbf{FRT-PAD} method consists of two branches, including a CNN based PA detector and an auxiliary branch. The CNN based PA detector disentangles disparities between bona fides and PAs by directly extracting features from image space. 
The auxiliary branch aims to extract face-related features from a model trained by face related tasks.
In such auxiliary branch, we firstly hierarchically obtain the task-specific features from multiple layers of the trained model.  
Then, we design a Cross-Modal Adapter based on GNN to adapt the features to PAD. The features from both branches are fused comprehensively for the final PAD. In the following sections, we will present the detailed discussion on the proposed method.

\subsection{Task Specific Features from Face Related Tasks}
As some face related tasks are trained from huge amount of faces, features extracted by such trained networks have better generalization capability. The hypothesis of this work is that face related tasks share some common patterns in face feature learning. For example, expression recognition requires the model to localize the action unit of the face, which also serves as a potential feature for PAD. 
Through transferring the knowledge contained in trained tasks, the dependence of PA detector on a large training data can be reduced. Hence features from face related tasks not only perform strong generalization in the trained task, but can also boost the performance of PA detector.
Let $x$ refers to a face in training set, and denote $FR(\cdot)$ as the network trained by face related tasks, e.g. face recognition, expression recognition and attribute editing. As shown in Fig.~\ref{fig:pipeline}, $FR(\cdot)$ embeds $x$ to a task specific feature $T =\{t_i| i \in [1,n]\}$, where $t_i$ refers to the feature map extracted from the $i$-th layer. As a multi-level representation of $x$, the features from different layer represent different properties of the face. To prevent from the loss of information, the proposed method regards such non-structure feature map as the input of the auxiliary branch. 

\begin{figure*}[h]
  \centering
   \includegraphics[width=0.88\linewidth]{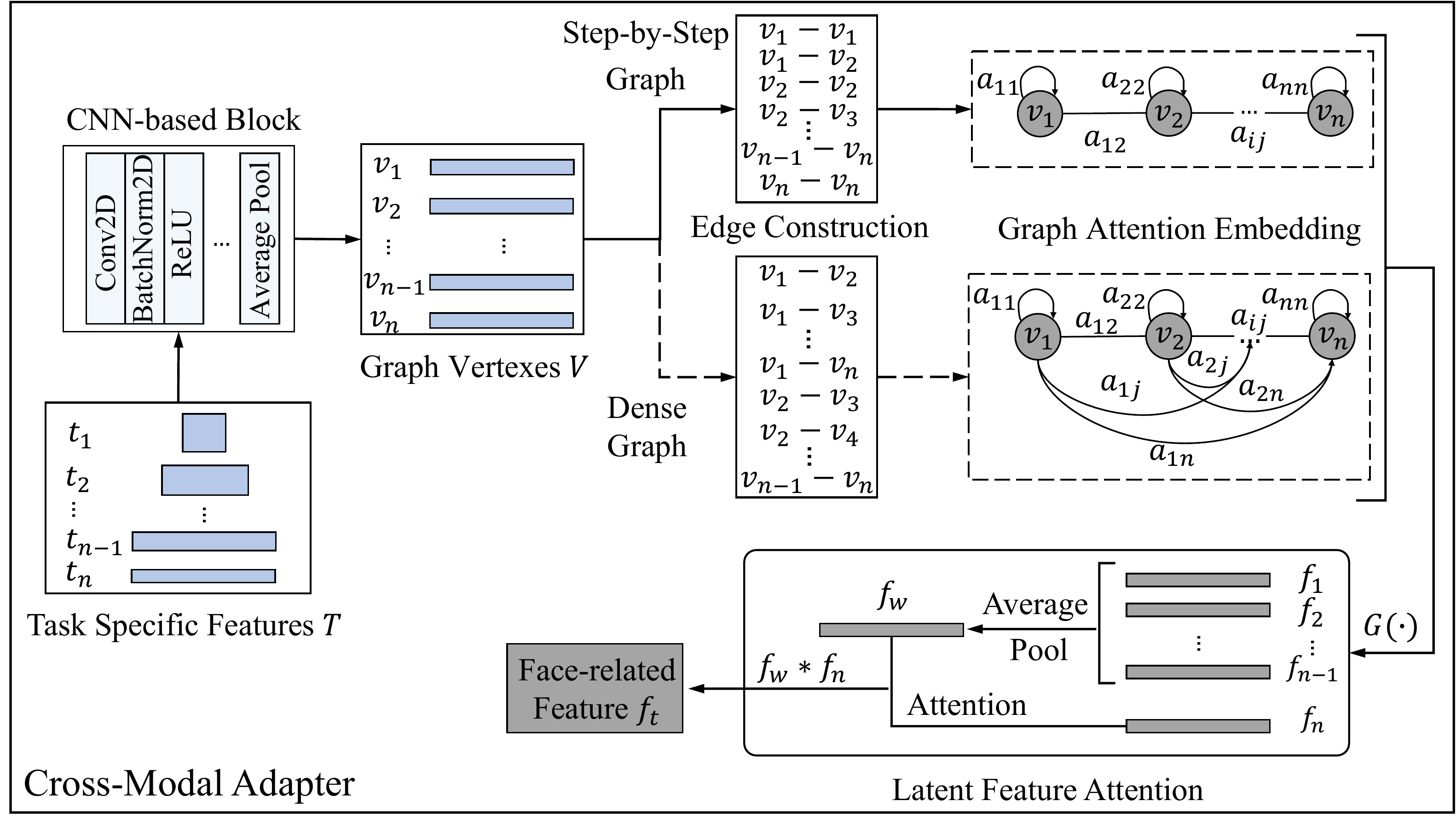}
   \caption{The pipeline of Cross-Modal Adapter. To transform the task specific features $t_i \in T$ to graph vertexes, $t_i$ are reconstructed to one-dimension vectors $v_i \in V$ using a CNN-based block. We design two different graphs, including Step-by-Step Graph and Dense Graph. In both graphs, attention mechanisms are used to specify the connection strength between different $v_i$. Through graph embedding $G(\cdot)$, $v_i$ can be re-mapped to $f_i$. Then, the latent feature $f_i$ is used to compute latent feature attention $f_w$ by an average pool operation. The final face-related feature $f_t$ is obtained by $f_w*f_n$. }
   \label{fig:gnn}
\end{figure*}
\subsection{Cross-Modal Representation Using GAT}
As $FR(\cdot)$ is trained by face related tasks, the extracted features $t_i$ might contain information unrelated to the face PAD task. To alleviate the potential negative influence of the irrelevant information, we propose a Cross-Modal Adapter to re-map them for PAD. 
Considering that task specific features are non-structural, a Graph Neural Network (GNN), denoted as $G(V, E)$, is employed to process $T$. $v_i \in V$ denotes vertex feature of graph. $E$ is the edge matrix of graph to connect neighboring vertexes given by:
\begin{equation}
E=\left\{\begin{matrix}
e_{ij}=1 & v_i\rightarrow v_j\\ 
e_{ij}=0 & v_i\overset{\times }{\rightarrow}v_j
\end{matrix}\right.
  \label{eq:edge}
\end{equation}
Given two graph vertexes $v_i$ and $v_j$, $e_{ij}=1$ presents an undirected edge existing between them.
To construct neighboring $v_i$ properly in $E$, the relationship among $v_i$ is needed to be exploited. 
Since $t_i$ is extracted by $FR(\cdot)$ step by step, $T$ can perform as a sequence. Based on such observation, we propose two potential graph structures, including Step-by-Step Graph and Dense Graph, to investigate the reasonable utilization of $T$. 

However, matrix $E$ can only reflect the connection of each $v_i$. 
The importance of different $v_i$ is unknown.
Thus, we adopt an attention mechanism in GNN (i.e. GAT) embedding to find out the contribution of neighboring vertex to the central vertex. More important vertex features will obtain larger weights. Formally, the relative weights of the connected graph vertexes can be computed by:
\begin{equation}
A = E * (q_1WV + (q_2WV)^T)
  \label{eq:att}
\end{equation}
where $Q=\{q_1, q_2\}$ is set as the shared attention parameter. A shared learnable weight matrix $W$ is set to achieve graph convolutional operation, which can map each vertex $v_i$ to a high-level feature. $A$ stands for the attention matrix of graph vertexes $V$. 
Then, we use softmax to normalize $a_{ij} \in A$ by following the rule of connection: 
\begin{equation}
A_{s}(i,j) = \frac{exp(a_{ij})}{\sum_{k \in [1,n]}exp(a_{ik}))}
  \label{eq:softmax}
\end{equation}
Given two connected points $v_i$ and $v_j$ , $A_{s}(i,j)$ and $A_{s}(j,i)$ measures the connection strength coefficient of them. 
For one-layer attention, the vertex features with attention weights can be obtained:  
\begin{equation}
V^{'}=A_{s}WV
  \label{eq:embed}
\end{equation}

As shown in Fig.~\ref{fig:gnn}, task specific features $t_i$ are firstly embedded into graph vertex features $v_i$ by a CNN-based block. Then, for Step-by-Step Graph, two graph vertexes $v_i$ and $v_{i+1}$ are sequentially connected by a single edge. Different graph vertexes are fully connected in Dense Graph. Through the multi-layer graph embedding $G(V^{'},E)$, the transformed features $f_i$ are obtained. 
We regard latent features as the attention weights to strengthen the representation of $f_n$. 
Through an average pooling operation along with $f_i,i \in [1,n-1]$, the latent feature attention $f_w$ can be computed. The final face-related feature $f_t$ is represented by $f_w*f_n$. 

\begin{algorithm*}[b!]  
\small
\caption{Presentation Attack Detection Using FRT-PAD}  
{\bf Input:} \\
\hspace*{0.12in} Training Set $\mathcal{X}_T$; CNN based PA detector $D(\cdot)$; Face recognition network $FR(\cdot)$; \\
\hspace*{0.12in} Classifier $C(\cdot)$; Graph embedding $G(\cdot)$; \\
{\bf Output:} \\ 
\hspace*{0.12in} Trained $D(\cdot)$, $G(\cdot)$ and $C(\cdot)$;
\begin{algorithmic}[1]
\State Fixed parameters of Trained $FR(\cdot)$;
\For{$x_j$ in $X_T$}
\State Extract \textbf{PAD feature} $f_p$ through $D(x_j)$;
\State Derive task specific features $t_i \in T$ from $FR(x_j)$;
\State Transform $t_i$ to vector $v_i \in V$ as graph vertexes; 
\State Construct edge matrix $E$;
\State Derive vertex features $V^{'}$ with attention weighs;
\State Extract transformed features $f_i$ through $G(V^{'},E)$; 
\State Obtain attention weights $f_w$ from $f_i, i\in[1,n-1]$; 
\State Calculate \textbf{face-related feature} $f_t$ by $f_w*f_n$; 
\State Derive \textbf{hierarchical feature} $f_h$ from $f_p$ and $f_t$; 
\State Predict the PAD result by $C(f_h)$;
\State Update $D(\cdot)$, $G(\cdot)$ and $C(\cdot)$by minimizing Eq.~\ref{eq:update};
\EndFor
\State Return $D(\cdot)$, $G(\cdot)$ and $C(\cdot)$; 
\end{algorithmic}
\label{alg_PAD} 
\end{algorithm*}
\subsection{Face-related-task based Presentation Attack Detection}
To adopt the re-mapped face-related feature in face PAD, the proposed method introduces a CNN based PA detector $D(\cdot)$ to learn the PAD feature $f_p$. 
Then, a hierarchical feature $f_h$ is derived by concatenating PAD feature $f_p$ and face-related feature $f_t$. 
Through $f_h$, classifier $C(\cdot)$ can effectively distinguish bona fides with PAs . 
In the training process, $D(\cdot)$, $G(\cdot)$ and $C(\cdot)$ are trained by a cross entropy as follows:
\begin{equation}
\mathcal{L}_{x_j \in \mathcal{X}_T}(x_j,y_j^{'}) = - \frac{1}{N}\sum_{j=1}^{N}[y_jlog(y_j^{'})+(1-y_j)log(1-y_j^{'})]
  \label{eq:update}
\end{equation}
where $(x_j, y_j), j\in [1,N]$ are the paired samples from training set $ \mathcal{X}_T$, and $y_j^{'}$ is the prediction result of $C(\cdot)$. 
For clarity, the proposed method is summarized in Algorithm~\ref{alg_PAD}.

\section{Experimental Results and Analysis}
In this section, we evaluate the performance of the proposed method by carrying experiments on the publicly-available datasets~\cite{boulkenafet2017oulu,zhang2012face,chingovska2012effectiveness,wen2015face}, First, the datasets and the corresponding implementation details are introduced. Then, we validate the effectiveness of the proposed method through analyzing the influences of each network component to PAD performance. Finally, to prove the superiority of our method, we compare the PAD performance of the proposed method with the state-of-the-art methods.

\subsection{Datasets and Implementation Details}
We use four public face anti-spoofing datasets, including OULU-NPU~\cite{boulkenafet2017oulu} (denoted as O), CASIA-FASD~\cite{zhang2012face} (denoted as C), Idiap Replay-Attack~\cite{chingovska2012effectiveness} (denoted as I) and MSU-MFSD~\cite{wen2015face} (denoted as M) to evaluate the effectiveness of our method. Existing methods were evaluated on the protocol~\cite{jia2020single}, denoted as Protocol-I. In this protocol, three of datasets are used as training set and the remaining one is used for test. However, in the reality, there are much more unseen PAs than the known ones in the training set. Using $3/4$ datasets to train model is not strict to the real application scenario. Thus, we design a different cross-dataset protocol (Protocol-II) to evaluate the generalization ability of our method. In detail, we only use two datasets from [O, M, C, I] to train model and the remaining two datasets to test. Due to the number of samples varies greatly among each dataset, some data divisions will be unreasonable for model training. To make the number of training set and test set as close as possible, we only divide the datasets into two groups, i.e. [O, M] and [C, I]. To reduce the influence caused by the background, resolution, and illustration, MTCNN algorithm~\cite{zhang2016joint} is used for face detection and alignment. All the detected faces are resized to (256, 256). ResNet18~\cite{he2016deep} is fine-tuned as the CNN based PA detector. Three network trained through the face related tasks are used to obtain task specific features. ResNet18 trained by face related task~\cite{deng2019arcface} and face expression recognition~\cite{wang2020suppressing} is set as the feature extractor. For face attribute editing task, the trained discriminator of StarGAN~\cite{choi2018stargan} is set to extract task specific features. 
In the graph embedding, a two-layer GAT with two-head attention mechanisms is adopted. In the training phase, the parameters of networks with face related tasks are fixed and the weight of task specific features are automatically determined. In terms of connection among vertexes, the attention mechanisms can specify the connection strength between them to show the most beneficial vertex. To evaluate the cross-modal adapter of our method, besides GAT, we adopt other deep learning methods, including ResNet18~\cite{he2016deep} and transformer~\cite{vaswani2017attention}, as the competing methods.

In summary, we train PA detector, classifier and cross-modal adapter by Adam with 1e-4 learning rate and 5e-5 weight decay. Batch size for training is 32. To validate the superiority of our FRT-PAD method, the state-of-the-art PAD methods, including DeepPixBiS~\cite{george2019deep}, SSDG-R~\cite{jia2020single}, CDC~\cite{yu2020searching}, and IF-OM~\cite{liu2021taming} are conducted in this paper. 
Following the work of~\cite{liu2021taming}, We use Half Total Error Rate (HTER), Area Under Curve (AUC) and Bona Fide Presentation Classification Error Rate (BPCER) when Attack Presentation Classification Error Rate ($APCER_{AP}$) is 1\% to evaluate the performance of PAD. 
This paper adopts the public platform pytorch for all experiments using a work station with CPUs of 2.8GHz, RAM of 512GB and GPUs of NVIDIA Tesla V100.

\subsection{Effectiveness Analysis of the Proposed Method}
\subsubsection{Face-related Features}

We perform the ablation study to quantify the influence of each component in our model for face PAD. 
First, to evaluate the effectiveness of face-related features, we test the performance of PAD with or without the face-related features. Table~\ref{tab:ablation_1} shows the results carried on the cross-dataset protocol. The baseline is set as the ResNet18 model pretrained from ImageNet. We use three face related tasks, including face recognition, face expression recognition, and face attribute editing to extract task specific features. Then, by a step-by-step GAT, we can respectively obtain three different face-related features ($F.R.$, $F.E.$, $F.A.$). 
Compared with the baseline, all three face-related features can improve the PAD performance. 
Specifically, $F.A.$ feature adapted from face attribute editing task improves the HTER of baseline from 26.90\% to 15.08\%. 
This indicates that face-related features are useful to face PAD.
As experiments are carried on the cross-dataset protocol, it also indicates that face-related features can improve generalization ability of face PAD. 
\begin{table}[h]
\centering
\caption{Performance of the Proposed Method with or without Face-related Features Obtained from Three Different Tasks. }
\resizebox{0.97\textwidth}{!}{%
\begin{tabular}{c|ccc|ccc}
\hline
\multirow{2}{*}{}      & \multicolumn{3}{c|}{{[}O.M{]} to {[}C,I{]}} & \multicolumn{3}{c}{{[}C,I{]} to {[}O,M{]}}                       \\ \cline{2-7} 
                       & HTER(\%)$\downarrow$           & AUC(\%)$\uparrow$          & BPCER(\%)$\downarrow$          & HTER(\%)$\downarrow$          & AUC(\%)$\uparrow$  & BPCER(\%)$\downarrow$   \\ \hline
Baseline           & 25.65   & 79.14  & 95.93   & 28.14  & 79.05 & 81.34   \\ \cline{1-7}
Baseline w/ $F.R.$ & 18.17   & 87.37   & 78.52  & 16.47  & 90.68 & 62.81   \\ \cline{1-7}
Baseline w/ $F.E.$ & 17.93   & 85.97  & 90.90    & 16.62  & 91.78 & 55.66   \\ \cline{1-7}
Baseline w/ $F.A.$ & \textbf{16.98} & \textbf{90.66} & \textbf{58.32} &\textbf{13.18} & \textbf{94.36} &\textbf{43.70} \\ \hline
\end{tabular}
\label{tab:ablation_1}
}
\end{table}

\begin{figure*}[t]
  \centering
   \includegraphics[width=1\linewidth]{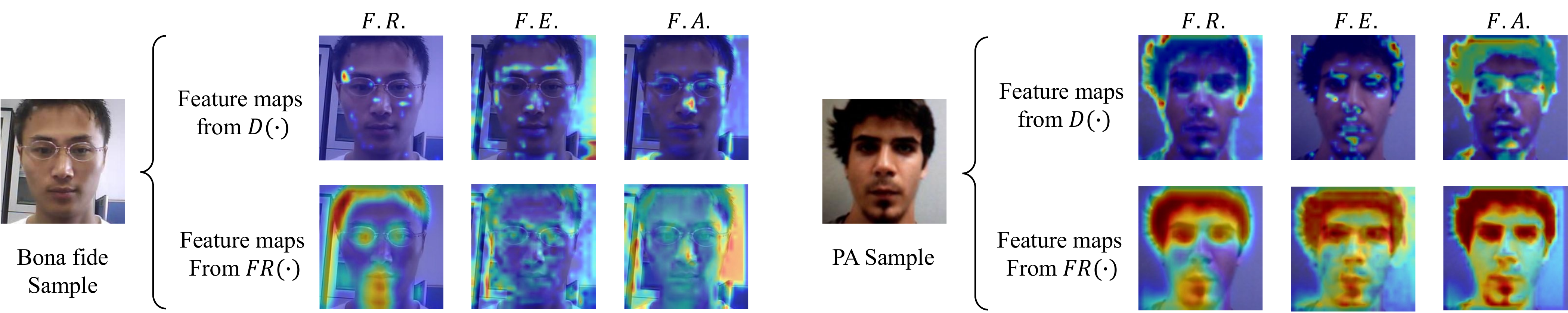}
   \caption{The visualization on CASIA-FASD using Grad-CAM. The first row for each sample shows the discriminative regions obtained from CNN-based PA detector $D(\cdot)$ and the second row for each sample illustrate the region localization extracted from network of face related tasks $FR(\cdot)$. }
   \label{fig:visual}
\end{figure*}

To further verify the effectiveness of face-related features, we adopt Grad-CAM~\cite{selvaraju2017grad} to visualize the discriminative regions from feature maps in our proposed FRT-PAD model. We compare the visualization results with three face-related features. Cross-Modal Adapters are set as the Step-by-Step Graphs. As shown in Fig.~\ref{fig:visual}, when using face-related features, the model can find discriminative features for both bona fide and PA samples. In the visualization region obtained form $FR(\cdot)$, the visualization shows that the hair, eyes, nose and mouth are important to distinguish live faces and spoofs. This further indicates the effectiveness of the face-related features. Typically, comparing with the visualization of $F.R.$ and $F.E.$ face-related features, $F.A.$ features can provide more effective region of face attributes. 

\subsubsection{Cross-Modal Adapter}
For each task specific feature $v_i$, we further compare two other different deep learning models with GAT, i.e. CNN based model and transformer model, to justify the effectiveness of the Cross-Modal Adapter.
In CNN based model, we use the same CNN-based block and latent feature attention in Fig.~\ref{fig:gnn} to obtain the face-related feature. In transformer based model, each $v_i$ is transformed to vector adopting CNN-based block in Fig.~\ref{fig:gnn} and encoded with position encoding module in~\cite{vaswani2017attention}. Then, we adopt six-layer transformer encoders with eight-head-attention to obtain the face-related feature. 
As CNN model and transformer model are sequential models, we only use the Step-by-Step Graph to ensure the fairness of the comparison. 
The PAD results in Table~\ref{tab:ablation_2} show that the performance of the proposed method with different deep learning models in Cross-Modal Adapter and face-related features is better than the baseline. 
\begin{table*}[h]
\caption{Performance of the Proposed Method Using Different Models in Cross-Modal Adapter for Three Face-related Features. }
\centering
\resizebox{0.99\textwidth}{!}{
\begin{tabular}{c|c|ccc|ccc}
\hline
\multirow{2}{*}{\begin{tabular}[c]{@{}c@{}}\textbf{Face-related}\\ \textbf{Features}\end{tabular}}   & \multirow{2}{*}{\begin{tabular}[c]{@{}c@{}}\textbf{Cross-Modal}\\ \textbf{Adapters}\end{tabular} }  & \multicolumn{3}{c|}{{[}O.M{]} to {[}C,I{]}} & \multicolumn{3}{c}{{[}C,I{]} to {[}O,M{]}} \\ \cline{3-8} 
                        &                & HTER(\%)$\downarrow$  & AUC(\%)$\uparrow$  & BPCER(\%)$\downarrow$ & HTER(\%)$\downarrow$ & AUC(\%)$\uparrow$ & BPCER(\%)$\downarrow$ \\\hline
                        
$\times$                &    $\times$    & 25.65    & 79.14    & 95.93   & 28.14   & 79.05   & 81.34   \\ \hline
\multirow{3}{*}{$F.R.$} &    CNN         & 19.89    & 87.56    & 78.71   & 16.70   & 91.59   & 52.60   \\ 
                        &    Transformer & 19.53    & 86.18    & 81.69   & 17.72   & 90.53   & 68.14   \\
                        &    GAT         & 18.17    & 87.37    & 78.52   & 16.47   & 90.68   & 62.81   \\ \hline\hline
\multirow{3}{*}{$F.E.$} &    CNN         & 21.58    & 86.47    & 65.95   & 17.94   & 89.93   & 57.86   \\
                        &    Transformer & 19.23    & 85.40    & 91.19    & 16.76   & 91.31   & 56.16   \\
                        &    GAT         & 17.93    & 85.97    & 90.90     & 16.62   & 91.78   & 55.66   \\ \hline\hline
\multirow{3}{*}{$F.A.$} &    CNN         & 18.05    & 86.34    & 89.93   & 16.55   & 90.50   & 60.16   \\
                        &    Transformer & 20.59    & 87.72    & 65.49   & 20.58   & 87.57   & 67.87   \\
                        &    GAT & \textbf{16.98}   & \textbf{90.66}    & \textbf{58.32}   &\textbf{13.18}   & \textbf{94.36}   &\textbf{43.70}  \\ \hline
\end{tabular}
}
\label{tab:ablation_2}
\end{table*}

\begin{figure*}[h!]
  \centering
   \includegraphics[width=0.99\linewidth]{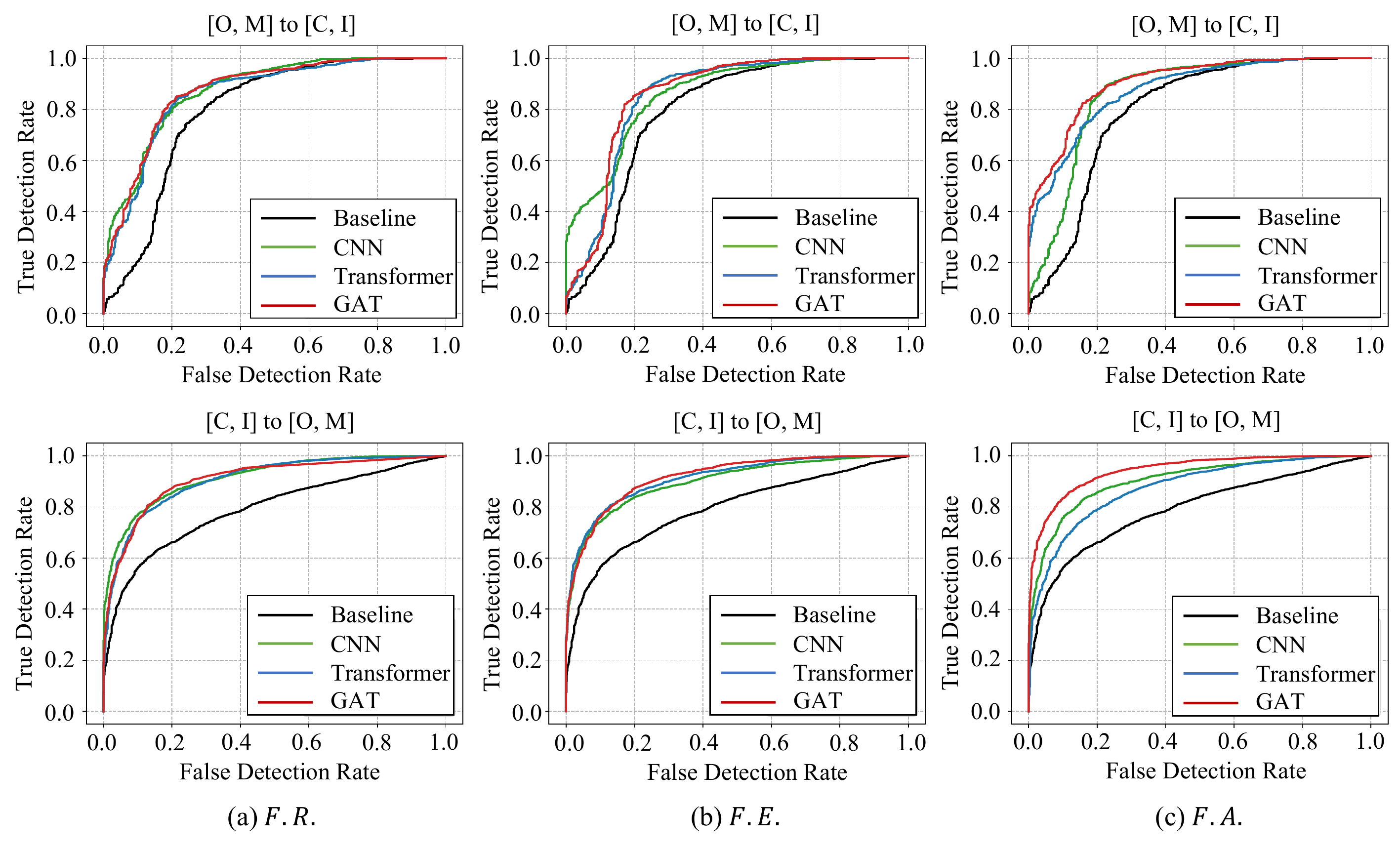}
   \caption{ROC curves for the ablation study when using different models in Cross-Modal Adapter. The experiments are under the cross-dataset setting (Protocol-II) and adopting three different face-related features, which are (a) $F.R.$, (b) $F.E.$ and (c) $F.A.$. Baseline (black line) in the figure represents the model without Cross-Modal Adapter. }
   \label{fig:roc}
\end{figure*}

Corresponding to Table~\ref{tab:ablation_2}, Fig.~\ref{fig:roc} presents the ROC (Receiver Operating Characteristic) curves of baseline and three face-related features when using different model in Cross-Modal Adapters. It can be seen that, for all face-related features, Cross-Modal Adapter using GAT model (with red lines in Fig.~\ref{fig:roc}) achieves a higher performance than CNN model and transformer model. This indicates that GAT model is more suitable for the Cross-Model Adapter. In particular, face-related feature $F.A.$ re-mapped from GAT can obtain the best results in [C, I] to [O, M] protocol, and achieve an HTER of 13.18\% and AUC of 94.36\%. These results can verify the contribution of Cross-Modal Adapter to improve the PAD performance and generalization.

\subsubsection{Different Backbones}
To further verify the effectiveness of face related task for PAD, we apply existing PAD methods, including  CDC~\cite{yu2020searching}, DeepPixBiS~\cite{george2019deep} and SSDG-R~\cite{jia2020single} as backbones. We compare the PAD results when above PAD methods with or without face-related features ($F.A.$) from face related tasks on Protocol-II. As shown in Table~\ref{tab:ablation_3}, the proposed face-related features can improve performance of all three methods. It indicates that the proposed method can be adopted to existing PAD methods and improve their generalization capability on unseen attacks. 
\begin{table*}[t]
\caption{Performance of Other PAD Backbones with or without Face-related Feature from FRT on Protocol-II. }
\centering
\resizebox{0.9\textwidth}{!}{
\begin{tabular}{c|ccc|ccc}
\hline
\multirow{2}{*}{}     & \multicolumn{3}{c|}{{[}O.M{]} to {[}C,I{]}} & \multicolumn{3}{c}{{[}C,I{]} to {[}O,M{]}} \\ \cline{2-7} 
                      & HTER(\%)$\downarrow$  & AUC(\%)$\uparrow$  & BPCER(\%)$\downarrow$  & HTER(\%)$\downarrow$  & AUC(\%)$\uparrow$  & BPCER(\%)$\downarrow$
                     \\ \hline
CDC~\cite{yu2020searching}               &  28.94   & 78.96  & 86.07     &23.30   &  83.42 & 74.17    \\
CDC w/FRT             & \textbf{25.95}     &\textbf{82.93}     &\textbf{81.82}      & \textbf{22.59}    & \textbf{84.06}    & \textbf{69.86}      \\ \hline
DeepPixBiS~\cite{george2019deep}        &  22.93   & 79.13  & 100.0         & 22.45  & 85.70 & 75.63     \\
DeepPixBiS w/ FRT     &  \textbf{21.93}   & \textbf{80.01}  & 100.0         & \textbf{19.44} & \textbf{88.89} &\textbf{65.46}       \\ \hline
SSDG-R~\cite{jia2020single}            &  20.92   &88.07 &90.28 &22.57 &85.61    &84.05    \\
SSDG-R w/FRT          &  \textbf{15.63}   & \textbf{91.77}  & \textbf{71.46}     &\textbf{18.68}   &\textbf{88.51}   &\textbf{54.92}    \\ \hline

\end{tabular}
}
\label{tab:ablation_3}
\end{table*}

\subsection{Comparison with other Methods}

To further verify the effectiveness of the proposed method, we compare it with the state-of-the-art methods in two protocols. Table~\ref{tab:comparison1} lists the comparison results in Protocol-I. Here, we give our results using the best model (i.e. adopting Step-by-Step Graph as the cross-modal Adapter of the $F.A.$ feature). 
It can be seen that, the proposed FRT-PAD method outperforms the state-of-the-art methods e.g. IF-OM and SSDG-R by the HTER. Specifically, in experiment [I, C, M] to O, our method can outperform both SSDG-R and IF-OM by average 2.60\% HTER.  
\begin{table*}[h]
\caption{Performance Comparison between the Proposed Method and the State-Of-The-Art Methods under the Cross-Dataset Setting. (Protocol-I).}
\Large
\resizebox{1.\textwidth}{!}{
\begin{tabular}{c|cc|cc|cc|cc}
\hline
                         & \multicolumn{2}{c|}{{[}O, C, I{]} to M} & \multicolumn{2}{c|}{{[}O, M, I{]} to C} & \multicolumn{2}{c|}{{[}O, C, M{]} to I} & \multicolumn{2}{c}{{[}I, C, M{]} to O} \\ \cline{2-9} 
\multirow{-2}{*}{Method} & HTER (\%)$\downarrow$          & AUC(\%)$\uparrow$            & HTER (\%)$\downarrow$          & AUC(\%)$\uparrow$             & HTER (\%)$\downarrow$          & AUC(\%)$\uparrow$             & HTER (\%)$\downarrow$          & AUC(\%)$\uparrow$       \\ \hline
MS-LBP~\cite{maatta2011face}  & 29.76              & 78.50              & 54.28              & 44.98              & 50.30              & 51.64              & 50.29              & 49.31             \\
Binary CNN~\cite{yang2014learn}  & 29.25              & 82.87              & 34.88              & 71.94              & 34.47              & 65.88              & 29.61              & 77.54            \\
IDA~\cite{wen2015face}    & 66.67              & 27.86              & 55.17              & 39.05              & 28.35              & 78.25              & 54.20              & 44.59            \\
Color Texture~\cite{boulkenafet2016face_} & 28.09              & 78.47              & 30.58              & 76.89              & 40.40              & 62.78              & 63.59              & 32.71          \\
LBP-TOP~\cite{de2014face}    & 36.90              & 70.80              & 33.52              & 73.15              & 29.14              & 71.69              & 30.17              & 77.61           \\
Auxiliary~\cite{liu2018learning}   & -                  & -                  & 28.40              & -                  & 27.60              & -                  & -                  & -                 \\
MADDG~\cite{shao2019multi}  & 17.69              & 88.06              & 24.50              & 84.51              & 22.19              & 84.99              & 27.89              & 80.02            \\ 
SSDG-R~\cite{jia2020single}      & 7.38               & 97.17              & \underline{10.44}     & \underline{95.94}     & \underline{11.71}              & \textbf{96.59}     & \underline{15.61}              & 91.54      \\
IF-OM~\cite{liu2021taming}  & \underline{7.14}      & \underline{97.09}     & 15.33     & 91.41     & 14.03     & 94.30     & 16.68     & \underline{91.85}    \\ \hline
Baseline                 & 13.10              & 92.76              & 16.44              & 91.25              & 24.58              & 79.50              & 22.31              & 85.65           \\ \hline
Ours: FRT-PAD  & \textbf{5.71}      & \textbf{97.21}     & \textbf{10.33}                & \textbf{96.73}              & \textbf{11.37}      & \underline{94.79}              & \textbf{13.55}     & \textbf{94.64}   \\ \hline
\end{tabular}
}
\label{tab:comparison1}
\end{table*}

Moreover, in more challenge Protocol-II (two datasets as training set and other two datasets as test set), we can obtain good results with both adapters (Dense Graph and Step-by-Step Graph) using the $F.A.$ feature.
As shown in Table~\ref{tab:comparison2}, the proposed method based on Step-by-Step Graph can outperform other methods by a large margin. Compared with CDC~\cite{yu2020searching}, our method can improve the HTER of PAD from 23.30\% to 13.18\% and AUC from 83.42\% to 94.36\%. 
These results indicate that the proposed method can generalize better than other PAD methods in both protocols, which further prove the superiority of our method.

\begin{table*}[t]
\caption{Performance Comparison between the Proposed Method and the State-Of-The-Art Methods under the Cross-Dataset Setting. (Protocol-II). }
\Large
\resizebox{\textwidth}{!}{%
\begin{tabular}{c|ccc|ccc}
\hline
                   & \multicolumn{3}{c|}{{[}O, M{]} to {[}C,I{]}}     & \multicolumn{3}{c}{{[}C, I{]} to {[}O, M{]}} \\ \cline{2-7} 
\multirow{-2}{*}{} & HTER(\%)$\downarrow$            & AUC(\%)$\uparrow$            & BPCER(\%)$\downarrow$            & HTER(\%)$\downarrow$            & AUC(\%)$\uparrow$            & BPCER(\%)$\downarrow$            \\ \hline
CDC~\cite{yu2020searching}                & 28.94          & 78.96          & 86.07          & 23.30          & 83.42          & 74.17          \\ \hline
DeepPixBiS~\cite{george2019deep}         & 22.93          & 79.13          & 100.0           & 22.45          & 85.70         & 75.63 \\ \hline
SSDG-R~\cite{jia2020single}             & 20.92          & 88.07          & 90.28           & 22.57          & 85.61          & 84.05          \\ \hline
IF-OM~\cite{liu2021taming}              & 18.96 & 89.48 & 69.52          & 18.60          & 89.76          & 69.70          \\ \hline
Baseline            & 25.65          & 79.14          & 95.93           & 28.14          & 79.05          & 81.34          \\ \hline

Ours: FRT-PAD w/ Dense Graph & 18.78    & 87.99     & 87.93            &16.30    &  92.32              &52.73            \\ \hline
Ours: FRT-PAD w/ Step-by-Step Graph & \textbf{16.98}         & \textbf{90.66}     & \textbf{58.32}            &\textbf{13.18}    &  \textbf{94.36}              &\textbf{43.70}             \\ \hline
\end{tabular}
}
\label{tab:comparison2}
\end{table*}

\section{Conclusion}
Existing face presentation attack detection methods cannot generalize well to unseen PAs, due to the highly dependence on the limited datasets. In this paper, to improve generalization ability of face PAD, we proposed a face PAD mechanism using feature-level prior knowledge from face related task in a common face system. By designing a Cross-Modal Adapter, features from other face related tasks can re-map to more effective features for PAD. Experimental results have shown the effectiveness of the proposed method. Compared with the state-of-the-art methods in existing dataset partition (i.e. Protocol-I), we can improve HTER to 5.71\%. Furthermore, when the dataset partition becomes more challenging (i.e. Protocol-II where more PAs are unseen to the model), our method largely improve the HTER to 13.18\%, which demonstrates the strong generalization ability of our method to handle unpredictable PAs.

\subsubsection{Acknowledgements} This work was supported in part by the National Natural Science Foundation of China (Grant 62076163 and Grant 91959108), and the Shenzhen Fundamental Research Fund (Grant JCYJ20190808163401646). Raghavendra Ramachandra is supported by SALT project funded by the Research Council of Norway. 

% and the Research Council of Norway (No. 321619 Project "OffPAD"). 
\bibliographystyle{splncs04}
\bibliography{egbib}

\newpage
\section*{APPENDIX}
\subsubsection{Different Baselines}

We set other baselines of PA detector, including VGG16 and swin-transformer-tiny to validate the proposed method in addition to ResNet18 used in the paper. As shown in Table~\ref{tab:1}, face attribute editing ($F.A.$) is adopted as the face related task (best case in the paper).
Our proposed method can improve performance over both baselines, which further verified the flexibility of our method against the weaker and stronger architectures.
\vspace{-1em}
\begin{table*}[h]
\caption{Performance of Other PAD Baselines with or without Face-related Feature from FRT on Protocol-II. }
\centering
\resizebox{\textwidth}{!}{
\begin{tabular}{c|ccc|ccc}
\hline
\multirow{2}{*}{}     & \multicolumn{3}{c|}{{[}O.M{]} to {[}C,I{]}} & \multicolumn{3}{c}{{[}C,I{]} to {[}O,M{]}} \\ \cline{2-7} 
                      & HTER(\%)$\downarrow$  & AUC(\%)$\uparrow$  & BPCER(\%)$\downarrow$  & HTER(\%)$\downarrow$  & AUC(\%)$\uparrow$  & BPCER(\%)$\downarrow$
                     \\ \hline
Vgg16               &  26.16 & 82.06 & 84.93    &26.50      & 81.81    &81.40     \\
Vgg16 w/FRT             & \textbf{19.18}     &\textbf{89.13}     &\textbf{62.22}      & \textbf{15.32}    & \textbf{92.24}    & \textbf{55.53}      \\ \hline
Swin-Transformer-tiny       &32.89       &66.95   & 100.00          &35.76    &67.27    &  84.19     \\
Swin-Transformer-tiny w/ FRT     &  \textbf{28.33}   & \textbf{78.31}  & \textbf{89.17}          & \textbf{31.40} & \textbf{72.45} &84.70       \\ \hline

\end{tabular}
}
\label{tab:1}
\end{table*}
\vspace{-2em} 

\subsubsection{Different Face Related Tasks and Models}
We applied different face related tasks and models in our FRT-PAD model. As shown in Table~\ref{tab:2}, we also adopted ResNet34 and ResNet50 models in face recognition task ($F.R.$). Both architectures of $F.R.$ brings similar improvement for PAD over the baseline detector (ResNet18), which proved that PA detector can be
benefited from the different architectures in the face related
tasks by following our solution.

On the other side, we applied other face related tasks in our method, i.e. face detection ($F.D.$) [1] and face localisation ($F.L.$) [2].
As listed in Table~\ref{tab:2}, consistent improvement can be observed, which further justifies the effectiveness of our method.
However, the performance of our method with ($F.D.$) and ($F.L.$) are lower than the case of $F.R.$, $F.E.$ and $F.A.$. One of the important reason is that ($F.D.$) and ($F.L.$) are trained to obtain location features rather than content features of faces. From this observation, the choice of the related face tasks is important. As an empirical result, ($F.A.$) performs the most suitable task to benefit PAD. 

\vspace{-1em}
\begin{table*}[h]
\caption{Performance of FRT Using Other Models and Tasks on Protocol-II. }
\centering
\resizebox{\textwidth}{!}{
\begin{tabular}{l|ccc|ccc}
\hline
\multirow{2}{*}{}     & \multicolumn{3}{c|}{{[}O.M{]} to {[}C,I{]}} & \multicolumn{3}{c}{{[}C,I{]} to {[}O,M{]}} \\ \cline{2-7} 
                      & HTER(\%)$\downarrow$  & AUC(\%)$\uparrow$  & BPCER(\%)$\downarrow$  & HTER(\%)$\downarrow$  & AUC(\%)$\uparrow$  & BPCER(\%)$\downarrow$
                     \\ \hline
Baseline (ResNet18)            & 25.65   & 79.14  & 95.93   & 28.14  & 79.05 & 81.34   \\ \cline{1-7}
Baseline w/ $F.R.$ (ResNet18)    &\textbf{18.17} & 87.37 & 78.52 & 16.47  & 90.68 & \textbf{62.81}     \\
Baseline w/ $F.R.$ (ResNet34)             & 19.86  &86.74   &74.51      & \textbf{16.16}     &\textbf{91.32}     &64.51      \\
Baseline w/ $F.R.$ (ResNet50)        &19.83  &\textbf{88.04}  &\textbf{73.05}          & 16.69   & 90.95  & 71.56     \\\hline
Baseline w/ $F.D.$ (ResNet50)             &21.47  &85.75  & 79.47   & 18.60       & 89.32    & 65.04      \\
Baseline w/ $F.L.$ (MobileNetV2)       &22.95  &82.98  & 82.52        & 19.51  &87.89  & 66.32      \\\hline

\end{tabular}
}
\label{tab:2}
\end{table*}

\vspace{-1em} 
\noindent[1] Deng J,et al. Retinaface: Single-shot multi-level face localisation in the wild. CVPR 2020. 

\noindent[2] Chen C. PyTorch Face Landmark: A Fast and Accurate Facial Landmark Detector.

\end{document}